# Mapping Road Lanes Using Laser Remission and Deep Neural Networks

Raphael V. Carneiro, Rafael C. Nascimento, Rânik Guidolini, Vinicius B. Cardoso,
Thiago Oliveira-Santos, Claudine Badue and Alberto F. De Souza, *Senior Member*, *IEEE*

Departamento de Informática
Universidade Federal do Espírito Santo
Vitória, ES, Brazil
{carneiro.raphael, rafaelnascimento, ranik, vinicius, todsantos, claudine, alberto}@lcad.inf.ufes.br

*Abstract*—We propose the use of deep neural networks (DNN) for solving the problem of inferring the position and relevant properties of lanes of urban roads with poor or absent horizontal signalization, in order to allow the operation of autonomous cars in such situations. We take a segmentation approach to the problem and use the Efficient Neural Network (ENet) DNN for segmenting LiDAR remission grid maps into road maps. We represent road maps using what we called *road grid maps*. Road grid maps are square matrixes and each element of these matrixes represents a small square region of real-world space. The value of each element is a code associated with the semantics of the road map. Our road grid maps contain all information about the roads' lanes required for building the Road Definition Data Files (RDDFs) that are necessary for the operation of our autonomous car, IARA (*Intelligent Autonomous Robotic Automobile*). We have built a dataset of tens of kilometers of manually marked road lanes and used part of it to train ENet to segment road grid maps from remission grid maps. After being trained, ENet achieved an average segmentation accuracy of 83.7%. We have tested the use of inferred road grid maps in the real world using IARA on a stretch of 3.7 km of urban roads and it has shown performance equivalent to that of the previous IARA's subsystem that uses a manually generated RDDF.

## I. INTRODUCTION

Autonomous cars need an internal model of the external world in order to properly operate in it. In urban roads, an autonomous car must stay within a lane in order to make space for other vehicles and, to achieve that, it must have some sort of internal map of the roads' lanes. Humans make use of the horizontal signalization of roads (road surface marking) to try and stay within lane while driving and there are many works in the literature about how to automatically detect relevant horizontal signalization for building ADAS systems [1]–[4] or relevant parts of autonomous cars systems [5]–[8]. However, sometimes the horizontal signalization is not in good conditions or even absent. This, although may make human driving more difficult, does not make human driving impossible, since humans can infer the position of road's lanes even in the absence of horizontal signalization. It may, nevertheless, make autonomous driving impossible if the autonomous car software system depends on horizontal road signalization.

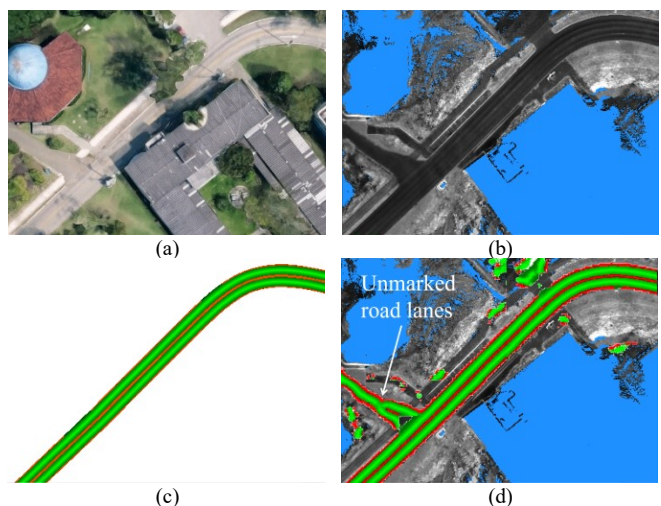

Fig. 1. (a) Google Maps image of a region of interest. (b) Remission grid map of (a); areas never touched by LiDAR rays are shown in blue. (c) Ground truth road map of (b). (d) DNN-generated road map; please note the inferred lanes that are unmarked in the ground truth.

In order to allow the operation of autonomous cars in urban roads with poor or absent horizontal signalization, we propose the use of a deep neural networks (DNN for short) for solving the problem of inferring the position and properties of the lanes of these roads. We take a segmentation approach to the problem and use the Efficient Neural Network (ENet) DNN [9] for segmenting LiDAR (Light Detection and Ranging) remission (intensity of the returned light pulses) grid maps (Fig. 1(b)) into road maps (Fig. 1(c)).

To train the DNN to segment road lanes from remission grid maps, we have built a dataset of tens of kilometers of manually marked road lanes (available at http://goo.gl/dmrebw). After being trained, the DNN achieved an average segmentation accuracy of 83.7% (% of correctly segmented cells of the remission grid maps). This value may seem small, but it actually reflects the inaccuracy of our ground truth (it is very hard to properly annotate road lanes manually, as we will discuss in

Research supported by Conselho Nacional de Desenvolvimento Científico e Tecnológico (CNPq), Brazil (grants 311120/2016-4 and 311504/2017-5), Vale S.A., Fundação de Amparo à Pesquisa do Espírito Santo (FAPES), Brazil (grant 75537958/16) and Coordenação de Aperfeiçoamento de Pessoal de Nível Superior (CAPES), Brazil.

Section IV.A) and the frequently appropriate inference made by the DNN, which is capable of finding existing lanes not marked in the ground truth (see Fig. 1(d)).

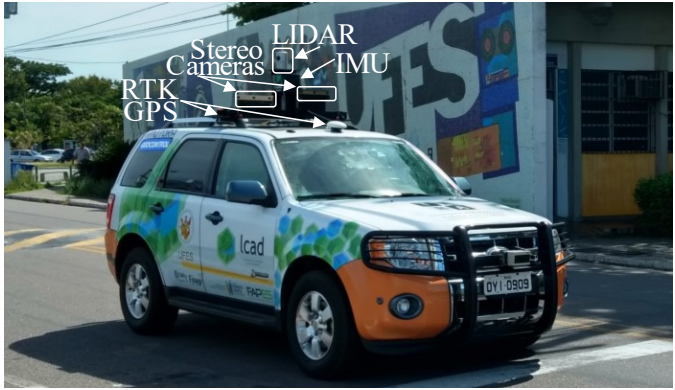

Fig. 2. Intelligent Autonomous Robotic Automobile (IARA). IARA uses several sensors to build an internal representation of the external world. A video of IARA's autonomous operation is available at https://goo.gl/RT1EBt.

We have tested the use of road lanes segmented by the ENet DNN in the real world using the *Intelligent Autonomous Robotic Automobile* (IARA, Fig. 2). IARA is an autonomous car we have developed at *Universidade Federal do Espírito Santo* (UFES), Brazil. It has all relevant sensors, hardware and software modules to allow autonomy (see Section III.A), but, prior to this work, IARA did not have an automatic mechanism for inferring the lanes of roads. It had to rely on precisely annotated information for autonomous operation, i.e. on Road Definition Data Files (RDDF [10]) – a sequence of world waypoints regularly spaced and associated annotations, such as road width, maximum speed, etc. However, now, thanks to the use of DNN for inferring road maps, IARA has a subsystem that can automatically extract a RDDF from the DNN-generated road maps (Fig. 3). We have tested this subsystem in IARA on a stretch of 3.7 km of urban roads and it has shown performance equivalent to that of the previous subsystem that use manually generated RDDFs.

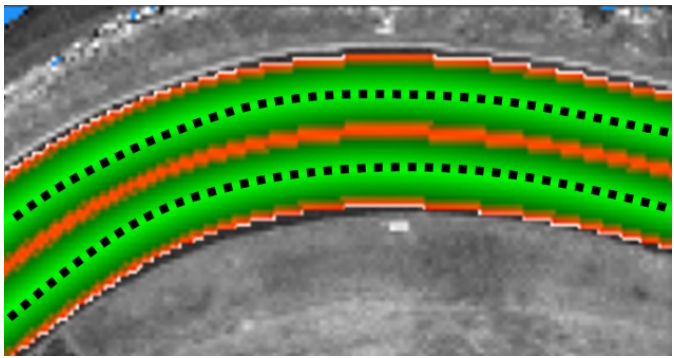

Fig. 3. Black dots correspond to RDDF waypoints automatically extracted from the road map, represented in shades of green and red.

## II. RELATED WORK

Diverse solutions were proposed for finding and classifying road lanes on images. In [11] Jung *et al.* present a method that uses aligned spatiotemporal images generated by accumulating the pixels on a scanline along the time axis. The aligned spatiotemporal image is binarized, and two dominant parallel straight lines resulting from the temporal consistency of lane width on a given scanline are detected using a Hough transform. The system outputs a cubic spline for each of the right and left lane (in the car's track), besides, each lane class. In [3] Berriel *et al.* propose a system that works on a temporal sequence of images applying an inverse perspective mapping and a particle filter for each image to track the lane along time. The system outputs a cubic spline for each of the right and left lane (in the car's track), besides, each lane class. Recently, Convolutional Neural Networks (CNNs) have surpassed image processing methods accuracy. In [12] Lee *et al.* built a dataset with 20,000 images with lanes labeled and trained a CNN to detect and track all lanes and road markings on the image at a pixel segmentation level. The system outputs a confidence map detecting and classifying the lanes on the image. These methods, however, are only used in the presence of road marks, do not produce a road map and were not tested on a real autonomous car. In addition, the method presented in [3] is strongly dependent on parameters.

A second set of solutions detects the road on the image on a pixel level segmentation. In [13], Wang *et al.* propose a road detection algorithm that provides a pixel-level confidence map of the road using likelihood. In [14], Laddha *et al.* propose an automatic dataset generation method, for pixel level road image annotations using OpenStreetMap, vehicle pose and camera parameters to train a road finding CNN. In [15], Teichmann *et al.* present an end-to-end approach to joint classification, detection and semantic segmentation via a unified architecture where the encoder is shared amongst the three tasks. The task of the encoder is to extract rich abstract features, the first decoder classifies the image in highway or minor road, the second decoder detect vehicles and the third decoder segment the image classifying road pixels. In [16], Wang *et al.* propose a Siamese fully convolutional network (FCN) to segment the road region. A semantic contour map (generated by a fast contour detection) and an RGB image are given as input for the Siamese FCN that give as output the per-pixel road regions. These methods, however, do not find the lanes on the road, do not produce a road map and were not tested on a real autonomous car.

There are other solutions that rely chiefly on LiDAR data for finding the road. Han *et al.* [17] propose a road detection method based on the fusion of LiDAR and image data. Firstly, LiDAR point clouds are projected into the monocular images by cross calibration and a joint bilateral filter to get high-resolution height. After that, pixels are classified using the scores computed from an Adaboost classifier based on the unary potential, the height value of each pixel, color information and pairwise potential. The system detects the road on both, image and LiDAR point cloud. Caltagirone *et al.* [18] propose a deep learning approach to carry out road detection using only LiDAR point cloud. The FCN receives as input LiDAR top-view images encoding several basic statistics such as mean elevation and density; this process reduced the problem to a single-scale problem. The FCN used is specifically designed for the task of pixel-wise semantic segmentation by combining a large receptive field with high-resolution feature maps. The system detects the road by classifying the LiDAR points that comprises the road. In [19], Caltagirone *et al.* developed a system to generate driving paths by integrating LiDAR point clouds, GPS-IMU information, and Google driving directions. The system is based on a FCN that jointly learns to carry out perception and path generation from real-world driving sequences and that is trained using automatically generated training examples. The system outputs a classification of the LiDAR points that belongs to the path. These methods, however, do not find the lanes on the road. Methods [17] and [18] find the road but [19] does not. None of these methods produce a road map, and they were not evaluated on a real autonomous car.

In [20], Hernández *et al.* propose a system based on the reflection of a 2D LiDAR to find the road and road lanes, but it does not produce a road map and was not tested on a real autonomous car. In [21], Kim and Yi also used 2D LiDAR reflectivity to find road lanes and build a lane map. Joshi and James, in [6], combine 3D LiDAR data and OpenStreetMap information to apply a particle filter-based approach, estimating the road lanes centerline and building a road map, but they do not estimate the road lane class. Gwon *et al.* [8] used 3D LiDAR intensity and GPS to extract lane marking position and type, estimate the road geometry and produce a road map representation with polynomial curves and evaluated on a real autonomous car. However, the aforementioned approaches employ classical machine learning methods based on human-defined feature extractors, which typically are outperformed by DNNs in visual recognition challenges such as Pascal VOC and ImageNet.

A DNN for Semantic Segmentation of Road Maps from Remission Grid Maps

We have used a DNN for the semantic segmentation of road maps from remission grid maps. In this work, we represent road maps using what we call *road grid maps*. The road grid maps we propose are square matrixes, $C = \{c_{0,0}, ..., c_{i,j}, ..., c_{s-1,s-1}\}$, where: $s \times s$ is the size of $C$ in cells; each element of $C$, $c_{i,j}$, represents a small square region of real-world space; and the value of each $c_{i,j}$ is a code associated with the semantics of the road grid map. We have used $s = 1,050$ and the size of the square region of real-world space represented by each $c_{i,j}$ is equal to 20 cm × 20 cm; so, a road grip map has a size of 210 m × 210 m. The possible values of $c_{i,j}$ are as follows (see Fig. 4):

- 0 = Off Lane;
- 1 = Solid Line marking;
- 2 = Broken Line marking;
- 3 = Solid Line marking (50% confidence);
- 4 = Broken Line marking (50% confidence);
- 5, 6, 7, ..., 16 = distance to the center of the lane (0, 1/22 of lane width, 2/22, ..., 11/22 of lane width, or 1/2 lane width – we have used a lane width of 3.2 m).

In the IARA Software System (see Section III.A), remission grid maps are square matrixes, $R = \{r_{0,0}, ..., r_{i,j}, ..., r_{s-1,s-1}\}$, where each element, $r_{i,j}$, represents a small square region of real-world space (we use 20 cm × 20 cm; the same size of the road grid map) and the value of each $r_{i,j}$ is the average LiDAR remission (intensity of the returned light pulse) of this region of real-world space observed during mapping, or –1 if it was never touched by any LiDAR ray (shown in blue in all figures of remission grid maps).

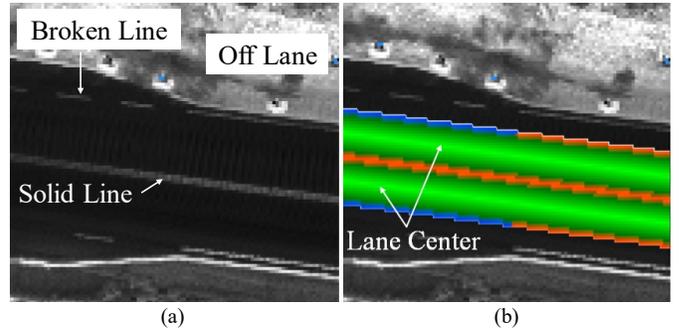

Fig. 4. (a) Crop of a remission grid map. (b) Same as (a) with superimposed road grid map ground truth. The value of each cell, $c_{i,j}$, of the road grid map ground truth represents different classes of features. $c_{i,j} = 0$ represents Off Lane (shown in (a)); $c_{i,j} = 1$ represents a Solid Line close to this lane boundary (shown in (a) and color coded in red in (b) – please note that we have chosen to mark lanes of fixed size in our ground truth); $c_{i,j} = 2$ represents a Broken Line close to this lane boundary (shown in (a), color coded in blue in (b)). A value of $c_{i,j}$ in the range from 5 to 16 represents different distances from this map cell to the center of the lane (color coded as different shades of green in (b)).

We have used the ENet DNN [9] for semantic segmentation of road grid maps from remission grid maps. ENet is a recently proposed encoder-decoder DNN architecture [22] [23]. An encoder-decoder DNN uses two separate neural network architectures combined together: an encoder and a decoder. In the ENet, the encoder is a convolution neural network that is trained to classify the input, while the decoder is used to upsample the output of the encoder. ENet is similar to SegNet [24] [25], but with a shorter decoder and several other modifications to allow faster operation without significant degradation of segmentation performance [9]. Also, ENet is trained in two stages. In the first stage, only the encoder is trained (receiving as input, in our case, a crop of a remission grid map, and as target the associated crop of the road grid map ground truth). Since de encoder reduces dimensionality by a factor of 8, it is necessary to perform an upsample of its output in order to compare to the ground truth. So, a temporary single deconvolution layer is connected at the output allowing the calculation of a softmax classifiers matrix with the size of the target ground truth (120 × 120 cells in our case). In the second stage, this temporary layer is removed and the decoder is connected at the output of the encoder (the output of the decoder is also a matrix of softmax classifiers with the size of the target ground truth). Then, the full network is trained.

During the test phase, the ENet receives as input a crop of the remission grid map and outputs the inferred road grid map in the form of a matrix of softmax classifiers: the class most active of each softmax classifier will be the value inferred for the corresponding $c_{i,j}$ of the road grid map.

III. THE USE OF ROAD GRID MAPS IN AUTONOMOUS CARS

The road grid map we have proposed has several relevant applications in the autonomous car area. It can be used, for example, to inform the distance from the autonomous car to the center of the nearest lane of a road, or to inform if the autonomous car can overtake another car through a specific side of the lane (using the broken or solid line information), or to build a graph of the road network mapped, etc. We use road grid maps to build RDDFs for the autonomous operation of IARA.

In the IARA Software System, RDDFs have a very simple text format and are composed of a sequence of waypoints (the distance between waypoints is about 0.5 m), recommended car orientations for each waypoint (yaw angle) and some road annotations, such as maximum velocity, presence of speed bumpers, crosswalks, etc. (each associated with a specific waypoint). To describe how we build and use RDDFs, it is relevant to first describe the IARA Software System.

### A. The IARA Software System

The software system that gives IARA its autonomy is composed of many modules (programs) that work together thanks to the Inter Process Communication library (IPC [26]). IPC implements the publish-subscribe paradigm and was incorporated into the Carnegie Mellon Robot Navigation Toolkit (CARMEN [27]). By the time IARA project started, in 2009, we chose CARMEN as our development framework because it was more stable and simpler than emerging platforms. Nevertheless, we plan to migrate IARA software to modern platforms such as ROS 2 in the future. We have extended CARMEN and created our own version (https://github.com/LCAD-UFES/carmen_lcad) that was used to develop IARA's autonomous system. When in autonomous mode, the IARA Software System is currently composed of 25 modules.

To describe how we build and use RDDFs from road grid maps, we briefly describe below the function of the IARA's modules that are relevant to their comprehension (see Fig. 5).

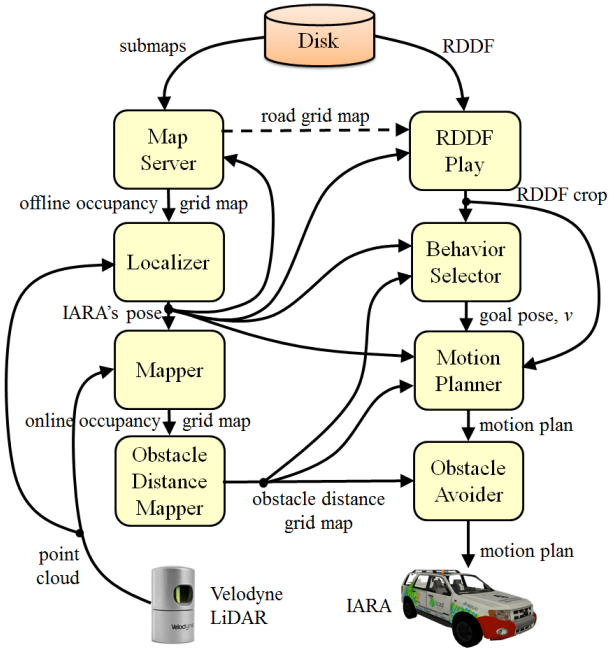

Fig. 5. IARA Software System simplified block diagram.

At startup, the IARA's Localizer module performs global localization [28] using GPS (GPS information is only used during startup). The Map Server module uses the initial localization to retrieve a remission map from the disk. A remission map of 210 m × 210 m is actually stored as nine submaps of 70 m × 70 m. The Map Server module fetches nine submaps from disk in such way as to have IARA in the central submap. As IARA moves, the Localizer module keeps track of its pose (we perform position tracking using a particle filter based on offline occupancy grid maps and LiDAR point clouds [29]) and publish it to the modules that have subscribed to it; at the same time, the Map Server fetches new submaps in order to keep IARA always within the central submap of a remission grid map (see video at http://goo.gl/dmrebw). Every time IARA crosses the limits of the central submap, new submaps are fetched from disk and a new remission grid map is built and published.

We compute remission grid maps using a technique similar to that of Levinson and Thrun [30] (see [31]). Actually, the IARA Software System produces and uses several other grid maps that have the same size of remission grid maps, and all these grid maps are published together with the remission grid maps (e.g. the offline occupancy grid maps mentioned above).

At IARA startup, the RDDF Play module retrieves from disk a RDDF that describes a mission the human operator wants IARA to perform. Typically, a mission comprises the autonomous navigation from IARA's current pose to the last waypoint in the RDDF. Given an IARA's pose, the RDDF Play crops a segment of the RDDF, composed of waypoints up to 75 m ahead and 25 m behind this pose, and publishes it. The Behavior Selector module subscribes to this message and, considering static obstacles, moving obstacles (other vehicles, typically), pedestrians, annotations in the RDDF crop, etc., publishes a goal (which is one of the RDDF's waypoints or annotations) and the desired velocity at this goal. The Motion Planner [32] subscribes to the RDDF crop, goal and goal velocity messages and plans the IARA movements from its current state (current pose, velocity and steering angle) to a state as close as possible to the goal pose and goal velocity. The Motion Planner computes and publishes optimum motion plans (20 times per second) that keep IARA as close as possible to the RDDF crop waypoints and at a safe distance from static and dynamic obstacles. The Obstacle Avoider [33] module subscribes to these motion plans and changes them, if needed, to avoid trajectories that may cause collisions.

Dynamic obstacles can be considered by the Motion Planner and Obstacle Avoider thanks to obstacle distance grid maps published by the Obstacle Distance Mapper module. Each cell of an obstacle distance grid map contains the distance and coordinates of the nearest (to this cell) obstacle. This module computes these maps from online occupancy grid maps [28] produced by the Mapper module from point clouds captured by a Velodyne HDL-32E LiDAR sensor.

### B. Computing RDDFs from Road Grid Maps

To compute the waypoints of a RDDF from road grid maps, we use the Algorithm 1 (when using RDDFs computed this way, the IARA Software System uses the dotted line in Fig. 5).

In Algorithm 1, lines 1 and 2 compute 150 waypoints ahead and 50 waypoints behind the current IARA's pose using the functions get_waypoints_ahead() and get_waypoints_behind(), respectively, comprising 75 m ahead and 25 m behind the

IARA's current pose. In line 3, these two sets of waypoints are smoothed by the function smooth_rddf_using_conjugate_gradient(). This smoothing is necessary because the waypoints computed in lines 1 and 2 are centered in map cells and their dimension (20 cm × 20 cm) is close to the separation between waypoints (0.5 m, or 50 cm).

**Algorithm 1:** compute_rddf_from_road_grid_map()
**Input:** *pose*, *road_grid_map*
**Output:** *RDDF_crop*
1: *rddf.wps_ahead* ← get_waypoints_ahead(*pose*, *road_grid_map*, *150*)
2: *rddf.wps_behind* ← get_waypoints_behind(*pose*, *road_grid_map*, *50*)
3: *RDDF_crop* ← smooth_rddf_using_conjugate_gradient(*rddf*)
4: **return** (*RDDF_crop*)

Algorithm 2 implements the function get_waypoints_ahead().

**Algorithm 2:** get_waypoints_ahead()
**Input:** *pose*, *road_grid_map*, *num_waypoints*
**Output:** *wps_ahead*
1: *next_pose* ← *pose*
2: **for** *i = 0* **to** *num_waypoints – 1*
3:   *wps_ahead[i]* ← get_lane_central_pose(*next_pose*, *road_grid_map*)
4:   *next_pose* ← add_distance_to_pose(*wps_ahead[i]*, *0.5*)
5:   *next_pose.yaw* ← atan2(*next_pose.y* – *wps_ahead[i].y*, *next_pose.x* – *wps_ahead[i].x*)
6: **return** (*wps_ahead*)

In Algorithm 2, line 3 computes a new waypoint for each value of the variable *i* using the function get_lane_central_pose(). This function receives as input *next_pose* = {*x*, *y*, *yaw*} and returns the cell in the road grid map (*road_grid_map*) that is the nearest to the center of the lane crossed by a line orthogonal to *next_pose* (see yellow line in Fig. 6). When *i* = 0, *next_pose* is equal to the IARA's current pose (line 1). So, get_lane_central_pose() finds the center of the lane nearest to the point where IARA is (its pose, see caption of Fig. 6). For the remainder values of *i*, this function returns waypoints along the center of the lane and 0.5 m spaced, thanks to the displacements added to *next_pose* in line 4.

The function get_waypoints_behind() is equivalent to get_waypoints_ahead() but finds points in the opposite direction. We implemented the function smooth_rddf_using_conjugate_gradient() modeling the smoothing optimization problem using the same objective function used by Dolgov *et al.* in [34], but considering only the term that measures the smoothness of the path ([34], equation 1).

IV. EXPERIMENTAL METHODOLOGY

To evaluate the use of road lanes segmented using the ENet DNN in the real world we have used the IARA autonomous car (Fig. 2).

We developed the hardware and software of IARA. The IARA's hardware is based on a Ford Escape Hybrid, which was adapted to: enable electronic actuation of the steering, throttle and brakes; reading the car odometry; and powering several high-performance computers and sensors. IARA has: one Velodyne HDL 32-E LiDAR; one Trimble dual RTK GPS; one Xsens MTi IMU; one Point Grey Bumblebee and one Point Grey Bumblebee XB3 stereo cameras; and two Dell Precision R5500 computers.

IARA's software is composed of many modules that cooperate via messages exchanged according to the publish-subscribe paradigm. These modules were implemented within the UFES version of the Carnegie Mellon Robot Navigation Toolkit (CARMEN [27]). All source code of IARA, including the source code developed to implement this work, is available at (https://github.com/LCAD-UFES/carmen_lcad).

*A. Datasets*

To train the DNN to segment road grid maps from remission grid maps, we have built a dataset of tens of kilometers of manually marked road lanes. We have made this dataset available at http://goo.gl/dmrebw). Part of the dataset pertains to the ring road of the *Universidade Federal do Espírito Santo* (UFES) (see Fig. 7), has an extension of 3.7 km and is named UFES dataset. The remaining part of the dataset pertains to highways (far away from UFES), has an extension of 32.4 km and is named Highway dataset.

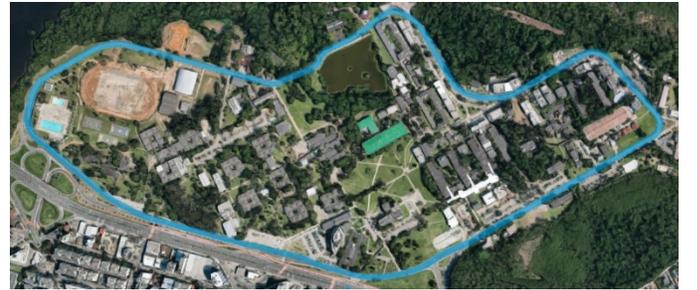

Fig. 7. The ring road of the *Universidade Federal do Espírito Santo* (UFES)

The UFES dataset comprises 658 crops (of 24 m × 24 m, i.e. 120 × 120 cells) of remission grid maps and corresponding road grid maps' ground truth taken along the UFES ring road about 5 m spaced. To build the ground truth, we have used the Inkscape (http://inkscape.org) vector graphics software. For that, we saved each remission grid line as a *png* image and used the following procedure we have developed:

1. Open a remission map image file using Inkscape and draw a polyline all the way along the center of each road lane.

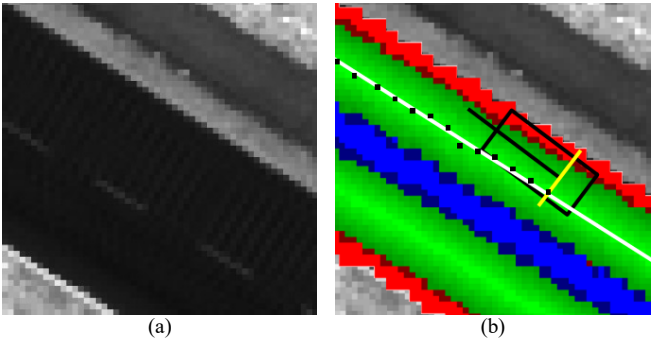

(a)                             (b)

Fig. 6. (a) Crop of a remission grid map. (b) Same as (a) with superimposed road grid map. The IARA is shown as a black rectangle – its pose is represented as the end of the black line segment inside the rectangle. The yellow line segment is the orthogonal line used for searching for the center of the lane. The black dots are the waypoints ahead (wps_ahead) found by get_lane_central_pose(), while the white curve is the final RDDF_crop, which is a smoothed version of wps_ahead and wps_behind produced by smooth_rddf_using_conjugate_gradient().

2. Place both the start and finish points out of the image limits, in order to provide a better fit for each polyline.
3. Place the points in such a way that the lanes' driving orientation matches the direction from the start point to the finish point.
4. If two polylines either cross, merge or fork, then make sure to draw the main polyline before the secondary.
5. Make each point of the polyline auto-smooth, in order to turn the polyline into a cubic Bezier curve.
6. Properly set each polyline stroke width to the width we have selected for all lanes (3.2 m, i.e. 16 cells).
7. Select each line stroke paint color according to the color code shown in http://goo.gl/dmrebw.

This procedure is very time consuming. So, to increase the size and diversity of the dataset, we have, for each pair of remission grid map and road grid map ground truth crops, produced a total of 168 pairs by: (i) rotating them by 24 different rotations (15 degrees apart), and (ii) translating them by 7 different translations (–1.5 m, –1.0 m, –0.5 m, 0.0 m, 0.5 m, 1.0 m, 1.5 m across the direction of the ring road at the position of the original pair of crops). The final UFES dataset has 110,544 (658 × 168) pairs of crops. We divided this dataset in two parts: (i) the UFES training dataset, with 88,368 pairs of crops; and (ii) the UFES test dataset, with 22,176 pairs of crops.

The Highway dataset comprises 3,556 crops (of 24 m × 24 m, i.e. 120 × 120 cells) of remission grid maps and corresponding road grid maps' ground truth taken along the highway about 12 m spaced. We used this dataset only for test; therefore, rotations and translations were not applied, and the dataset was not divided. We used the Highway dataset to evaluate our model generalization in a different context, not seen during training of the ENet.

*B. ENet DNN*

We have used a slightly modified version of ENet, that we made available at https://goo.gl/BzoRPK. Basically, we changed it to receive grayscale *png* 120 × 120-pixel images representing the remission grid maps as input, and to receive 120 × 120-pixel images representing road grid maps ground truth as target. To train our ENet, we have used the UFES training dataset, and to test it we have used the UFES test dataset and the Highway dataset.

The training parameters for the encoder network were: batch size of 16; number of batches trained 16,569 (3 epochs × 88,368 samples / 16 samples per batch = 16,569 batches); initial learning rate of 0.005, reduced by a factor of 10 every 4,142 batches; solver type ADAM, with first momentum equal to 0.9, second equal to 0.999, and weight decay equal to 0.0002. The training parameters for the full network (encoder-decoder) were the same.

The metric used to evaluate the performance of ENet is the class average accuracy, measured as the number of correct inferred classes divided by the total number of map cells evaluated. Since selecting a good metric for a specific segmentation problem is not trivial, this choice was guided by its simplicity and ease of calculation. Additionally, in order to evaluate relevant aspects for our application, the following qualitative criteria of navigation performance were considered: (i) keeping the car in the lane center, and (ii) avoiding abrupt steering movements.

V. EXPERIMENTS

Fig. 8 presents the full ENet (encoder-decoder) training experimental results. In Fig. 8, the *x* axis shows the number of batches trained, while the *y* axis shows the network accuracy. As the graph of Fig. 8 shows, the accuracy increases with the number of training batches, but reaches a plateau at about 2,000 batches. It starts increasing again when the learning rate decreases to 0.0005 and reaches another plateau at about 5,000 batches. The other reductions of the learning rate do not appear to increase the accuracy further, which reaches a plateau of about 78% during the training.

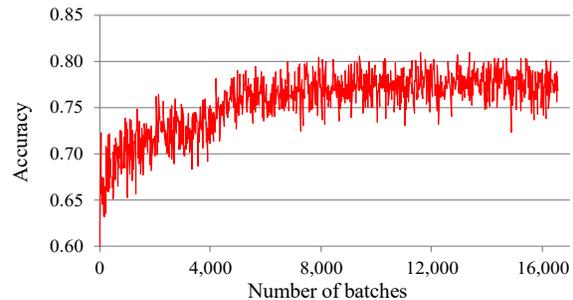

Fig. 8. Full ENet encoder-decoder accuracy achieved during training.

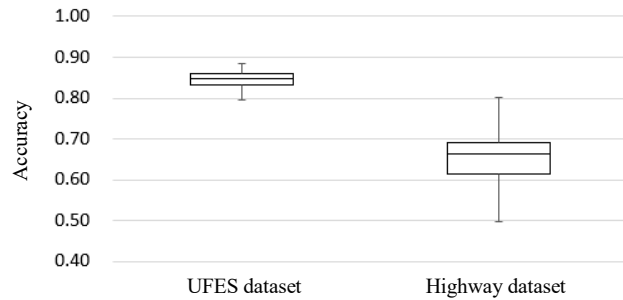

Fig. 9. Box plot of accuracy achieved during testing.

As Fig. 9 shows, the average accuracy of ENet on the UFES test dataset was 83.7%; a value higher than that observed during training. This happened because of the SpatialDropout [35] operation used during training for improving generalization. During test, SpatialDropout is removed, which increases the measured performance. Fig. 10 shows an example of the ENet inference on a segment of the UFES test dataset. Fig. 10 (a) shows the input remission grid map and Fig. 10 (b) shows the same remission grid map with the superimposed inferred road grid map.

In the test of ENet on the Highway dataset, an average accuracy of 64.1% was achieved, as shown in Fig. 9. Fig. 11 presents an example of test of ENet on the Highway dataset. As this figure shows, despite the lower accuracy measure, the DNN was able to make a fair extraction of the road grid map from the remission grid map of a region never seen during training and coming from very different surroundings.

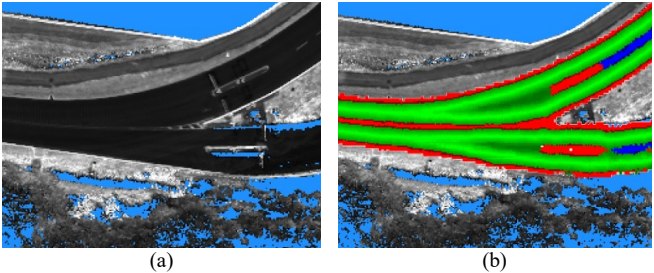

Fig. 10. Test of inference on the ring road of UFES. (a) Crop of a remission grid map. (b) Same as (a) with superimposed road grid map inferred by ENet with 83% of class average accuracy.

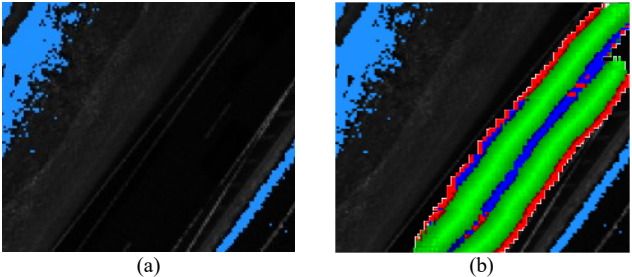

Fig. 11. Test of inference on the Highway dataset. (a) Crop of a remission grid map. (b) Same as (a) with superimposed inferred road grid map with 64% of class average accuracy.

To test the capability of the proposed system to compute proper RDDFs from road grid maps, firstly, we have set IARA to drive autonomously using a RDDF extracted from the manually annotated road grid map ground truth of the ring road of UFES. In this experiment, IARA have shown an equivalent autonomous navigation performance than when using the precisely annotated RDDF (the RDDF employed by the previous subsystem, used before the implementation described in this paper). No human intervention was needed. Then, ENet was used to generate the road grid map of the ring road of UFES and we set IARA to drive autonomously using the RDDF extracted from this road grid map. In this experiment, IARA had a navigation performance equivalent to that of the first experiment described above. Again, no human intervention was needed. A video comparing these two experiments is available at http://goo.gl/dmrebw.

One important question to be answered is: what is a good evaluation measure for this specific semantic segmentation problem? This question is not trivial as shown in Csurka *et al*. [36]. Different segmentation algorithms might be optimal for different measures, so the evaluation measure must be linked to the success of the end application. In our case, the end application is the IARA navigation subsystem, so a good road grid map is one which assures the shape and the center of each lane.

An important point to mention about the new IARA Software System, that now uses road grid maps, is that, currently, the only high-level decisions that IARA needs to make for traveling to a destination are, basically, to decide (i) if it should change lanes (for overtaking another vehicle, for example) or (ii) which direction it has to choose at intersections (if it should take right, or left, or if just continue straight ahead, see, for example, Fig. 10). This makes path planning easier and one may simply use online tools, such as OpenStreetMap, Google Maps or Waze, for this level of path planning, as humans do when traveling to a destination for the first time.

## VI. CONCLUSIONS AND FUTURE WORK

In this paper, we presented an approach for allowing the operation of autonomous cars in urban roads with poor or absent horizontal signalization. For that, we have used the ENet DNN for semantic segmentation of road maps from remission grid maps captured via LiDAR remission mapping. We represent road maps using what we called road grid maps, which are square matrixes where each element represents a small square region of real-world space and codes the semantics of the road grid map at this region. Using simple algorithms, we can extract RDDFs from road grid maps and use them to guide autonomous cars from their initial position to a desired destination.

We examined experimentally the semantic segmentation performance of ENet using a dataset we have built by manually marking tens of kilometers of roads. Our results have shown that ENet can achieve an accuracy of 83.7%. This value may seem small, but it actually reflects the inaccuracy of our ground truth (it is very hard to mark lanes of roads manually) and the frequently appropriate inference made by the DNN, which is capable of finding existing lanes not marked in the ground truth.

We have tested our approach in the IARA autonomous car on a stretch of 3.7 km of urban roads and it has shown performance equivalent to that of a system that uses manually generated RDDFs. This result suggests that a DNN segmentation accuracy of 83.7% in the dataset we have used in our experiments is enough for proper autonomous operation with our approach.

As future work, we plan to better tune the DNN architecture we have used, try other architectures and train and test them with our complete dataset. *Transfer Learning* might be a potential solution to avoid labeling each and every new road condition. In addition, we plan to examine mechanisms for extracting full road networks from road grid maps and store them as graphs, in order to allow inferences of the kind necessary for autonomous taxi services. We will also examine the use of online tools, such as OpenStreetMap, Google Maps or Waze, as alternatives for making online high-level path planning decisions in the context of our road grid maps.

## ACKNOWLEDGMENT

The authors would like to thank the NVIDIA Corporation for their kind donation of GPUs. We also thank the students Gabriel Hendrix and Gabriella A. Ridolphi for their voluntary effort in manually marking the road lanes in the ground truth datasets.